\pdfoutput=1

\documentclass[11pt]{article}

\usepackage{acl}

\usepackage{times}
\usepackage{latexsym}

\usepackage{amsmath}
\usepackage{booktabs}
\usepackage{graphicx}
\usepackage{todonotes}
\usepackage{tabularx}
\usepackage{xcolor}
\definecolor{correct}{HTML}{009900}

\usepackage[T1]{fontenc}

\usepackage[utf8]{inputenc}

\usepackage{microtype}

\usepackage{inconsolata}

\title{Select and Summarize: Scene Saliency for Movie Script Summarization}

\author{Rohit Saxena \qquad Frank Keller \\
 Institute for Language, Cognition and Computation\\
 School of Informatics, University of Edinburg \\
 10 Crichton Street, Edinburgh EH8 9AB \\
  \texttt{rohit.saxena@ed.ac.uk} \quad \texttt{keller@inf.ed.ac.uk}}

\begin{document}
\maketitle
\begin{abstract}
Abstractive summarization for long-form narrative texts such as movie scripts is challenging due to the computational and memory constraints of current language models. A movie script typically comprises a large number of scenes; however, only a fraction of these scenes are salient, i.e., important for understanding the overall narrative. The salience of a scene can be operationalized by considering it as salient if it is mentioned in the summary. Automatically identifying salient scenes is difficult due to the lack of suitable datasets. In this work, we introduce a scene saliency dataset that consists of human-annotated salient scenes for 100 movies. We propose a two-stage abstractive summarization approach which first identifies the salient scenes in script and then generates a summary using only those scenes. Using QA-based evaluation, we show that our model outperforms previous state-of-the-art summarization methods and reflects the information content of a movie more accurately than a model that takes the whole movie script as input.\footnote{Our dataset and code is released at \href{ https://github.com/saxenarohit/select_summ}{ https://github.com/saxenarohit/select\_summ}.}
\end{abstract}

\section{Introduction}
Abstractive summarization is the process of reducing an information source to its most important content by generating a coherent summary. Previous work has primarily focused on news \citep{cheng-lapata-2016-neural,gehrmann-etal-2018-bottom}, meetings \citep{zhong-etal-2021-qmsum}, and dialogues \citep{zhong2022dialoglm,zhu-etal-2021-mediasum}, but there is limited prior work on summarizing long-form narrative texts such as movie scripts \citep{gorinski-lapata-2015-movie,chen-etal-2022-summscreen}. 

Long-form narrative summarization poses challenges to large language models \citep{Beltagy2020Longformer,pmlr-v119-zhang20ae,huang-etal-2021-efficient} both in terms of memory complexity and in terms of attending to salient information in the text. Large language models perform poorly for long sequence lengths in zero-shot settings compared to fine-tuned models \citep{shaham2023zeroscrolls}. Recently, \citet{liu2023lost} showed that the performance of these models degrades when the relevant information is present in the middle of a long document. With an average length of 110 pages, movie scripts are therefore challenging to summarize. 

Several methods have previously relied on content selection for summarization to reduce the input size by either performing content selection implicitly using neural network attention \citep{chen-bansal-2018-fast,you-etal-2019-improving,zhong-etal-2021-qmsum} or explicitly \citep{ladhak-etal-2020-exploring,manakul-gales-2021-long,zhang-etal-2022-summn} by aligning the source document with the summary using metrics such as ROUGE \citep{lin-2004-rouge}. Unlike for news articles, the implicit attention-based method is problematic for movie scripts, as current methods cannot reliably process text of such length. On the other hand, current explicit methods are neither optimized nor evaluated for content selection using gold-standard labels. In addition, considering the large number of sentences in movies that contain repeated mentions of characters and locations, a method based on a lexical overlap metric such as ROUGE creates many false positives. Crucially, all these methods use source--summary alignment as an auxiliary task without actually optimizing or evaluating this task.

For news summarization, \citet{ernst-etal-2021-summary} created crowd-sourced development and test sets for the evaluation of proposition-level alignment. However, news texts differ from movie scripts both in length and in terms of the rigid inverted pyramid structure that is typical for news articles. For movie scripts, \citet{mirza-etal-2021-alignarr} proposed a specialized alignment method which they evaluated on a set of 10 movies. However, they do not perform movie script summarization.

Movie scripts are structured in terms of scenes, where each scene describes a distinct plot element and hapening at a fixed place and time, and involving a fixed set of characters. It therefore makes sense to formalize movie summarization as the identification of the most salient scenes from a movie, followed by the generation of an abstractive summary of those scenes \citep{gorinski-lapata-2015-movie}. Hence we define movie scene saliency based on whether the scene is mentioned in the summary i.e., if the scene is mentioned in the summary, it is considered salient. Using scene saliency for summarization is therefore a method of explicit content selection.

In this paper, we first introduce MENSA, a \textbf{M}ovie Sc\textbf{EN}e \textbf{SA}liency dataset that includes human annotation of salient scenes in movie scripts. Our annotators manually align Wikipedia summary sentences with movie scenes for 100 movies. We use these gold-standard annotations to evaluate existing explicit alignment methods. We then propose a supervised scene saliency classification model to identify salient scenes given a movie script. Specifically, we use the alignment method that performs best on the gold-standard data to generate silver-standard labels on a larger dataset, on which we then train a sequence classification model using scene embeddings to identify salient scenes. We then fine-tune a pre-trained language model using only the salient scenes to generate movie summaries. This model achieves new state-of-the-art summarization results as measured by ROUGE and BERTScore \citep{bert-score}. In addition to that, we evaluate the generated summaries using a question-answer-based metric \citep{deutsch-etal-2021-towards} and show that summaries generated using only the salient scenes outperform those generated using the entire movie script or baseline models. 

\section{Related Work}
\subsection{Long-form Summarization}
Summarization of long-form documents has been studied across various domains, such as news articles \citep{10.1145/3404835.3462846}, books \citep{kryscinski2021booksum}, dialogues \citep{zhong2022dialoglm}, meetings \citep{zhong-etal-2021-qmsum}, and scientific publications \citep{cohan-etal-2018-discourse}. To handle and process the long documents, many efficient transformer variants have been proposed \citep{NEURIPS2020_c8512d14,pmlr-v119-zhang20ae,huang-etal-2021-efficient}.
Similarly, work such as Longformer \citep{Beltagy2020Longformer} uses local and global attention in transformers \citep{NIPS2017_3f5ee243} to process long inputs. However, given that movie scripts are particularly long (see Table~\ref{tab:stats}), these models still have a limited capacity due to memory and time complexity, and need to truncate movie scripts based on the maximum sequence length supported by the model. 

Over the past decade, numerous approaches movie summarization have been proposed. 
\citet{gorinski-lapata-2018-whats,gorinski-lapata-2015-movie} generate movie overviews using a graph-based model and create movie script summaries based on progression, diversity, and importance. In contrast, the aim of our work is to find salient scenes and use these for summarization. \citet{papalampidi-etal-2019-movie,papalampidi2021movie} summarize movie scripts by identifying turning points, important narrative events. In contrast, our approach is based on salient scenes and does not assume a rigid narrative structure. Recently, \citet{agarwal-etal-2022-creativesumm} proposed a shared task for script summarization; the best model \citep{pu-etal-2022-two} used a heuristic approach to truncate the script. 

\subsection{\scalebox{0.97}{Summarization based on Content Selection}}
Several methods \citep{ladhak-etal-2020-exploring,manakul-gales-2021-long,liu-etal-2022-long} have leveraged content selection for summarization. \citet{chen-bansal-2018-fast} and \citet{zhang-etal-2022-summn} generate silver standard labels through greedy alignment of the source document sentences with summary sentences. However, these methods do not explicitly evaluate alignments. Moreover, movie scripts consist of a large number of sentences with the same characters and location names, which can generate many false positives in greedy alignment. 
We collect gold-standard saliency labels to compare and evaluate alignment methods. \citet{mirza-etal-2021-alignarr} proposed a movie script alignment method for summaries but do not actually propose a summarization model. Recent work \citep{dou-etal-2021-gsum,wang2022salience} has employed neural network attention for the summarization of short documents. However, movie scripts are challenging for attention-based methods, given their length. 

\section{MENSA: Movie Scene Saliency Dataset}
We define the saliency of a movie scene based on the mention of the scene in a user-written summary of the movie. If the scene appears in the summary, then it is considered salient for understanding the narrative of the movie. By aligning summary sentences to movie scenes, we identify salient scenes and later use them for movie summarization.

The MENSA dataset consists of the scripts of 100 movies and respective Wikipedia plot summaries annotated with gold-standard sentence-to-scene alignment. We selected 80 movies randomly from ScriptBase \citep{gorinski-lapata-2015-movie} and added 20 recently released, manually corrected movie scripts, which all had Wikipedia summaries.

Both MENSA and ScriptBase datasets are movie scripts datasets and differ from other dialogue/narrative datasets such as SummScreenFD \citep{chen-etal-2022-summscreen}, the ForeverDreaming subset of the SummScreen dataset as used in the \textsc{Scrolls} benchmark \citep{shaham-etal-2022-scrolls}. SummScreenFD is dataset of TV show episodes and consists of crowd-sourced transcripts and recaps. In contrast, the movie scripts in our dataset were written by screenwriters and the summaries were curated by Wikipedia. It is important to note that movies and TV shows have different storytelling structures, number of acts, and length. SummScreenFD has shorter input texts and summaries compared to movie scripts as shown in Table~\ref{tab:dataset_comp}.
\begin{table}[htbp]
\centering
\scalebox{0.9}{%
\begin{tabular}{@{}lr@{}}
\toprule
Total number of movies & 100         \\
Total number of scenes & 16,208         \\
Total number of summary sentences  & 3,295      \\
Sentence-Scene alignment pairs & 6,063 \\
Total number of salient scenes & 5,365 \\
\bottomrule
\end{tabular}}
\caption{Statistics of the MENSA dataset.}
\label{tab:stats}
\end{table}
\vspace{-1ex}
\begin{table}[htbp]
\centering
\scalebox{0.9}{%
\begin{tabular}{@{}lcc@{}}
\toprule
&SummScrFD & MENSA\\
\midrule
Mean script length & 7,605 & 35,926\\
Mean summary length & 113 & 860\\
\bottomrule
\end{tabular}}
\caption{Statistics of the length of the script and summary in the SummScreenFD and MENSA datasets.}
\label{tab:dataset_comp}
\end{table}
\subsection{Annotation Scheme}
Formally, let $M$ denote a movie script consisting of a sequence of scenes $M = \{S_1,S_2,...,S_N\}$ and let $D$ denote the Wikipedia plot summary consisting of a sequence of sentences $D = \{s_1,s_2,...,s_T\}$. The aim is to annotate and select a subset of salient scenes $M'$ such that $M' \subset M$ and $|M'|\ll|M|$, where for every scene in $M'$ there exist one or more aligned sentences in~$D$.

To manually align the summary sentences for 100 movies, we recruited five in-house annotators. They received detailed annotation instructions and were trained by the authors until they were able to perform the alignment task reliably. To analyze inter-annotator agreement, 15 movies were selected randomly and triple-annotated by the annotators. The remaining 85 movies were single annotated, similar to the annotation process used by \citet{papalampidi-etal-2019-movie}, to reduce the cost of annotation. As annotating and aligning a full-length movie script with its summary is a difficult task, we provided a default alignment to annotators generated by the alignment model of \citet{mirza-etal-2021-alignarr}. For every summary sentence, annotators first verified the default alignment with movie script scenes. If the alignment was only partially correct or missing, they corrected the alignment by adding or removing scenes for a given sentence using a web-based tool.
We assume that each sentence can be aligned to one or more scenes and vice versa. In Table~\ref{tab:stats}, we present statistics of the scripts and summaries in the MENSA dataset.

To evaluate the quality of the annotations collected, we computed inter-annotator agreement on the triple annotated movies using three metrics: (a)~Exact Match Agreement ($EMA$), (b)~Partial Agreement ($PA$), and (c)~Mean Annotation Distance ($D$). These measures were used for a similar annotation task by \citet{papalampidi-etal-2019-movie}.\footnote{We renamed total agreement in \citet{papalampidi-etal-2019-movie} to EMA for clarity.} EMA is the ratio of the intersection of the scenes that the three annotators exactly agree upon for a given summary sentence, which is averaged over all sentences in the summary (Jaccard Similarity) and computed as follows:
\begin{equation}
EMA = \frac{1}{T_M} \sum_{s=1}^{T_M} \frac{| A_{s} \cap B_{s} \cap C_{s}|}{| A_{s} \cup B_{s} \cup C_{s} |}
\end{equation} 
where $T_M$ is the total number of sentences in all the summaries, and $A_{s}$, $B_{s}$, and $C_{s}$ are the indices of the scenes selected for sentence $s$ by the three annotators.

Partial agreement ($PA$) is the ratio where there is an overlap of at least one scene among the annotators and is given as follows:
\begin{equation}
PA = \frac{1}{T_M} \sum_{s=1}^{T_M} [ A_{s} \cap B_{s} \cap C_{s} \neq \emptyset]
\end{equation} 
Annotation distance ($d$) for a summary sentence $s$ between two annotators is defined as the minimum overlap distance and is computed as follows:
\begin{equation}
d_s[A,B] =   \min_{\forall i \in A_s, \forall j \in B_s} |i-j|
\end{equation}
where $A_{s}$ and $B_{s}$ are the indices of the scenes selected for a sentence $s$ by the two annotators.
The mean annotation distance ($D$) between the three annotators is defined as the maximum pairwise overlapping annotation distance averaged for three annotators across all sentences:
\begin{equation}
D =  \frac{1}{T_M} \sum_{s=1}^{T_M}\max(d_s[A,B,C]) 
\end{equation}
where $d_s[A, B, C]$ is the pairwise annotation distance between three annotators and $T_M$ is the total number of sentences in all the summaries.

EMA and PA between our annotators was $52.80\%$ and $81.63\%$, respectively. The PA indicates that for every sentence in the summaries, there is a high overlap of at least one scene. This is consistent with the low mean annotation distance of $1.21$, which indicates that on average the distance between the annotations is around one scene. The EMA shows that for more than half of the sentences, there is an exact match in scene-to-sentence alignment among the annotators.

\begin{table}[htbp]
\centering
\scalebox{0.9}{%
\begin{tabular}{@{}lrrr@{}}
\toprule
 \textbf{Alignment Method} & \textbf{P}        & \textbf{R}       & \textbf{F1}       \\ \midrule
\citet{chen-bansal-2018-fast}     & 52.59 & 51.38 & 51.67          \\
\citet{zhang-etal-2022-summn}      & 50.35 & 53.42 & 50.15         \\ 
\citet{mirza-etal-2021-alignarr} & \textbf{84.42} & \textbf{68.53} & \textbf{73.55} \\
\bottomrule
\end{tabular}}
\caption{Comparing alignment performance for different alignment methods on the gold-standard set.}
\label{tab:alignment_exp}
\end{table}

\begin{figure*}
  \centering
\vspace{-2ex}
\includegraphics[width={0.9\textwidth}]{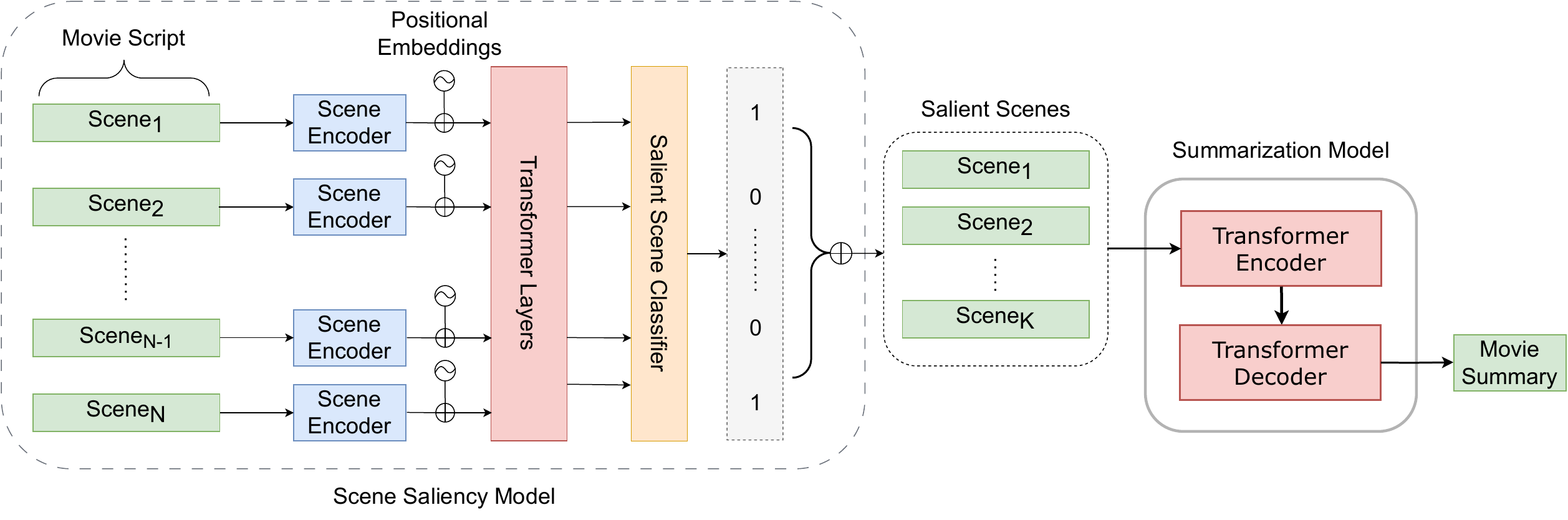}
\vspace{-2ex}
  \caption{The architecture of the scene saliency detection and summarization models. The models are trained in a pipeline where salient scene detection is trained separately.}
  \label{fig:model}
\end{figure*}

\subsection{\scalebox{0.95}{Evaluation of Automatic Alignment Methods}}

Since it is too expensive and time-consuming to collect gold-standard scene saliency labels for the whole of Scriptbase \citep{gorinski-lapata-2015-movie}, we generate silver-standard labels to train a model for scene saliency classification. Based on our definition of scene saliency above, silver-standard labels for scene saliency can be generated by aligning movie scenes with summary sentences. 

Alignment between the source document segments and the summary sentences has been previously proposed for news summarization \cite{chen-bansal-2018-fast,zhang-etal-2022-summn} and narrative text \cite{mirza-etal-2021-alignarr}. Using our gold-standard labels, we investigate which of these approaches yields better alignment between movie scripts and summaries and therefore should be used to generate silver-standard labels for scene saliency. 

\citet{chen-bansal-2018-fast} used ROUGE-L to align a summary sentence to the most similar source document sentence. 
In our case, we transformed these source document (movie script) sentence-level alignments to scene-level alignments such that if the scene contains the aligned sentence, the scene will be aligned to the summary sentence. 
\citet{zhang-etal-2022-summn} used a greedy algorithm for aligning the document segment and the summary sentences. For each segment, the sentences are aligned based on the gain in ROUGE-1 score. In our case, movie scenes are considered as source document segments. \citet{mirza-etal-2021-alignarr} proposed an alignment method specifically for movie scripts using semantic similarity combined with Integer Linear Programming (ILP) to align movie script scenes to summary sentences. 

We present the results of applying these three approaches on our gold-standard MENSA dataset in Table~\ref{tab:alignment_exp}. We report macro-averaged precision ($P$), recall ($R$), and $F1$ score. The \citet{mirza-etal-2021-alignarr} method performs significantly better than the ROUGE-based methods, possibly as it was specifically proposed to align movie scenes and summary sentences.\footnote{It was also used to generate the default alignment that our human annotators had to correct, which biases our evaluation towards the method of \citet{mirza-etal-2021-alignarr}. However, our results are still a good measure of how many errors human annotators find in the alignment generated by this method.} We therefore used this alignment method to generate silver-standard scene saliency labels for the complete Scriptbase corpus. 

Our dataset can be used in the future to evaluate content selection strategies in long documents. The gold-standard salient scenes can also be used to evaluate extractive summarization methods.

We now introduce our Select and Summarize (\textsc{Select \& Summ}) model, which first uses a classification model (Section~\ref{scene_sal_model}) to predict the salient scenes and then utilizes only the salient scenes to generate a movie summary using a pre-trained abstractive summarization model (Section~\ref{summar_model}). These models are trained in a two-stage pipeline. 

\section{Scene Saliency Classification Model}\label{scene_sal_model}

Using the set of generated silver-standard labels for scene saliency, we train a neural network-based classification model to predict scene saliency. We formulate this task as a sequence labeling task where the model takes a sequence of scenes $M = \{S_1,S_2,...,S_N\}$ as input and predicts a sequence of binary labels $Y = \{y_1,y_2,...,y_N\}$ denoting whether a scene is salient. 

The model consists of two components, as shown in Figure~\ref{fig:model}. The first component is a scene encoder which computes scene representations by concatenating the sentences in the scene and encodes them using a pre-trained language model. Next, to learn contextual scene representation across the whole movie, we further encode the scene embeddings generated by the scene encoder using a transformer \citep{NIPS2017_3f5ee243} block ($L$ layers stacked), with unmasked self-attention initialized with random weights \citep{liu-lapata-2019-text}. To preserve the sequence of the scenes, we add positional encodings to scene representations obtained from the first component. The final contextualized representation of the scenes is then used to classify whether scenes are salient or not. The model is trained for binary sequence labeling using the binary cross-entropy loss.

\subsection{Dataset}\label{dataset}

To train the saliency model, we used the ScriptBase corpus \citep{gorinski-lapata-2015-movie} that contains preprocessed scripts of movies with Wikipedia summaries. We removed the movies used in our gold-standard MENSA dataset from Scriptbase from the training set. This resulted in a training set containing 824 movie scripts, for which we generated silver-standard scene saliency labels using the model of \citet{mirza-etal-2021-alignarr}, as previously discussed. We randomly split our gold-standard scene saliency dataset of 100 movies, using half of it for validation and the other half for testing.

\subsection{Baselines}

\textbf{Majority Class:} We used predicting the majority class as a simple baseline for classification. The dataset is highly imbalanced, with non-salient being the majority class.\\
\textbf{Unsupervised TextRank:} We used an extension of TextRank \citep{mihalcea-tarau-2004-textrank,zheng-lapata-2019-sentence}, a graph-based algorithm which is used for unsupervised extractive summarization. Similar to \citet{papalampidi-etal-2020-screenplay}, instead of a sentence-based graph we constructed a movie script graph such that nodes in the graph correspond to the scenes in the movie~$M$. The edge $e_{ij}$ between any two scene nodes $S_i$ and $S_j$ represents their similarity, with the edge weight being the similarity score. 
The centrality of a node $S_i$ measures the importance of that node (in our case, the node represents the scene) and is computed as follows:
\begin{equation*}
\mathit{centrality}(S_i)= \lambda_1 \sum_{j<i}^{}e_{ij} + \lambda_2 \sum_{j>i}^{}e_{ij}
\end{equation*}
where $\lambda_1$ and $\lambda_2$ are weights for forward-looking (edges to following scene nodes) and backward-looking (edges to preceding scene nodes) and sum to one. In our experiments, we represent the scene by computing a scene representation using a pre-trained language model (see below). We compute the weight of the edge between two nodes using the cosine similarity between the scene representations and select top-K nodes as the salient scenes based on their centrality score.\\
\textbf{Supervised Bi-LSTM:} For a supervised baseline, we used a bi-directional LSTM (Bi-LSTM) to learn contextual representation for the classification of scene saliency. Again, we computed scene representations by concatenating the sentences and encoding them using a pre-trained language model.

Note that the alignment model of \citet{mirza-etal-2021-alignarr} cannot be used as a baseline for saliency classification: it requires summaries to align to movie scripts at test time. In a summarization scenario, no summaries are available at test time. 

\subsection{Implementation Details}\label{impl_sal}
Our Scene Saliency Model and baseline models employed RoBERTa-large as the pre-trained scene encoder. Representation of a scene is computed using the first token's last hidden state of the model. The movie encoder transformer block has 10 layers with 16 heads and a feedforward hidden size of 2048. As the binary scene labels in the dataset are highly imbalanced, we used weighted binary cross entropy. We employed AdamW with $\beta_1= 0.9$, $\beta_2 = 0.999$ as our optimizer. The learning rate was fixed at 5e-5. For the baseline Text\-Rank model, we performed a grid search for hyperparameters and used $\lambda_1 = 0.7$, $\lambda_2=0.3$, and $K=15\%$ of move length. For Bi-LSTM, we used hidden dimension of size 512 followed by a fully connected layer. 
\begin{figure}
  \centering
\includegraphics[width=0.7\columnwidth]{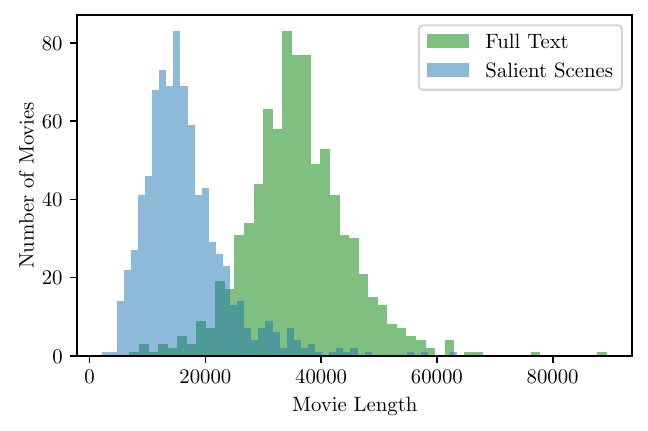}
\vspace{-2ex}
  \caption{Distribution of movie length from the training set for full text and only the salient scenes.}
  \label{fig:length_dist}
\end{figure}

\begin{table}[htbp]
\centering
\scalebox{0.9}{%
\begin{tabular}{@{}lrrr@{}}
\toprule
\textbf{Method} & \textbf{P}        & \textbf{R}       & \textbf{F1}       \\ \midrule
Majority Class     & 31.99 & 50.00 & 38.70          \\
TextRank      & 56.15 & 53.57 & 50.55        \\ 
Bi-LSTM      & 64.39 & 61.85 & 61.17        \\ 
Scene Saliency Model & \textbf{68.38} & \textbf{68.13} & \textbf{68.01} \\
\bottomrule
\end{tabular}}
\caption{Comparing saliency classification performance for different classification models and baseline; macro-averaged precision ($P$), recall ($R$), and F1.}
\label{tab:scene_sal}
\end{table}
\subsection{Results}

The results of our saliency classification model and the baselines are summarized in Table~\ref{tab:scene_sal}. We report macro-averaged precision (P), recall (R), and F1 score for each model, as the labels are highly imbalanced given that only a limited number of scenes in each movie are salient. Our model outperforms the baselines and achieves 68.38, 68.13, and 68.01 on precision, recall, and F1. The results show that the majority baseline performance is equivalent to random guessing (macro-average). The unsupervised TextRank model has higher precision, recall, and F1 than the majority baseline, which indicates that it is able to correctly predict some scenes as salient based on the centrality score. Also, the high value of $\lambda_1$ (see Section~\ref{impl_sal}) signifies that the backward-looking context is more important than forward-looking context for computing scene importance. The transformer-based scene saliency model achieves better performance than other baselines, indicating the effectiveness of transformer layers in learning the context across scene representations, which is helpful in classifying scene saliency. We also found that a higher number of layers worked better for the transformer, which indicates that more layers help in capturing complex relationships in the input. See Appendix~\ref{sec:robustness} for $k$-fold cross-validation on the test set. 
\begin{table*}[tb]
\centering
\scalebox{0.9}{%
\begin{tabular}{@{}lrrrrrr@{}} 
\toprule            
 & \textbf{R-1} & \textbf{R-2} & \textbf{R-L} & \textbf{BS\textsubscript{p}} & \textbf{BS\textsubscript{r}} & \textbf{BS\textsubscript{f1}} \\ 
\midrule
Lead-512       & 10.30 & 1.22 & 9.73  & 49.89 & 43.88 & 46.68     \\
Lead-1024       & 17.69 & 2.10 & 16.78  & 49.43 & 46.86 & 48.10        \\
CreativeSumm \citep{agarwal-etal-2022-creativesumm}* & 14.92  & 1.46  & 13.73  & 42.98  & 42.38   & 42.58 \\
\midrule
FLAN-T5-XXL (Zero-Shot)  & 10.82 & 1.13 & 10.54 & 46.22 & 35.81 & 39.48\\
Vicuna-13b-1.5 (Zero-Shot) & 16.26 & 3.67 & 15.39 & 53.22 & 48.60  & 50.80  \\
GPT-3.5 Turbo (Zero-Shot) & 22.53 & 3.69 & 20.49 & 49.71 & 46.49 & 48.01\\
FLAN-UL2 (Zero-Shot) & 25.86 & 5.42 & 24.71 & 51.40 & 48.79 & 50.71\\

\midrule
Random Selection (LED)      & 41.28 & 8.99 & 39.81 & 54.13 & 53.96 & 54.04\\
\textsc{Summ$^N$} Multi Stage \citep{zhang-etal-2022-summn} & 22.65 & 3.01 & 22.42 & 40.87 & 38.60 & 39.68\\
Unlimiformer \citep{bertsch2023unlimiformer}	& 31.12 &	4.15 &	30.25 & 41.74 & 51.55 & 46.08\\
Two-Stage Heuristic \citep{pu-etal-2022-two} & 46.89  & 11.11  & 44.65  & 56.39  & 56.30   & 56.34 \\
\midrule
Full Text (Pegasus-X) & 46.20 & 10.58 & 45.35 & 57.10 & 57.01 & 57.05  \\
\textsc{Select \& Summ} (Pegasus-X) & 48.29 & 11.40  & 46.62 & 57.39 & 57.20 & 57.31\\
Full Text (LED) & 46.15  & 10.62  & 44.46 & 56.32 & 55.77  & 56.04  \\
\textsc{Select \& Summ} (LED) & \textbf{49.98}  & \textbf{12.11}  & \textbf{47.95} & \textbf{57.64} & \textbf{57.29} & \textbf{57.46}   \\
\bottomrule
\end{tabular}%
}
\caption{Results of our model Select and Summarize (\textsc{Select \& Summ}) compared with other summarization models.
*Denotes model results from the paper of the shared task.}
\label{tab:summ_result}
\end{table*}

\section{Summarization Using Salient Scenes}\label{summar_model}
We now investigate the benefit of using only salient scenes for the abstractive summarization of movie scripts. We formulate this task as a sequence-to-sequence generation problem. Formally, given a movie with a set of salient scenes $M = \{S_1, S_2,...,S_K\}$, the goal is to generate a target summary $S = \{s_1, s_2,...,s_m\}$. As the input length of the salient scenes is still quite large as shown in Figure~\ref{fig:length_dist}, we use a Longformer Encoder-Decoder (LED) architecture \citep{Beltagy2020Longformer}. To handle long input sequences, LED uses efficient local attention with global attention for the encoder. The decoder then uses the full self-attention to the encoded tokens and to previously decoded locations to generate the summary.

\subsection{Dataset}
We used the same dataset and split as in Section~\ref{dataset}, now with Wikipedia plot summaries as output for movie script summarization. However, instead of using the whole movie script, we utilize the output of our scene saliency model and input only the salient scenes when we generate movie summaries.

\subsection{Baselines}
We compare the proposed model with various baselines. \textbf{Lead-N} simply outputs the first $N$ tokens of the movie script as the summary of the movie. 
We varied $N$ to understand the impact of summary length on performance and report results on \textbf{Lead-512} and \textbf{Lead-1024}. \textbf{FLAN-T5-XXL} \citep{chung2022scaling}, \textbf{FLAN-UL2} \citep{wei2022finetuned}, \textbf{Vicuna-13b-1.5} \citep{zheng2023judging} which is fine-tuned on Llama-2 \citep{touvron2023llama}, and \textbf{GPT-3.5-Turbo}\footnote{We used model gpt-3.5-turbo-1106 which has context length of 16K tokens.} \cite{NEURIPS2020_1457c0d6} are instruction-tuned large language models (LLMs) which were used in zero-shot setting. \textsc{\textbf{SUMM\textsuperscript{N}}} \citep{zhang-etal-2022-summn} is a multi-stage summarization framework for long input dialogues and documents. \textbf{Unlimiformer} \cite{bertsch2023unlimiformer} uses retrieval-based attention mechanism for long document summarization.
\textbf{Two-Stage Heuristic} \cite{pu-etal-2022-two} is a two-stage movie script summarization model which first selects the essential sentences based on heuristics and then summarizes the text using LED with efficient fine-tuning. \textbf{Random Selection} randomly selects salient scenes for summarization. \textbf{Full Text} takes the full movie script as input (no content selection) and truncates the text based on model input length.

\subsection{Implementation Details}\label{impl_summ}
We experimented with two pre-trained models LED and Pegasus-X as base models for summarization which were fined-tuned on the Scriptbase corpus (see Section~\ref{dataset}). Each input sequence for the movie is truncated to 16,384 tokens (including special tokens) to fit into the maximum input length of the model. We experimented with both the base and large variants of these models and found that the large models performed better and used them in our experiments. We used AdamW as an optimizer ($\beta_1 = 0.9$, $\beta_2 = 0.99$) with a learning rate of 5e-5. We used a linear warmup strategy with 512 warmup steps. We trained the models to 60 epochs and used the checkpoint with the best validation score. We used a beam size of five for decoding and generating the summary. We also created a random selection baseline by selecting a random $k$\% of scenes and using those to generate a summary. We report the best result for random selection, which was obtained for $k=25$ and LED. All the baseline models are fully trained on our dataset using the best configuration from the papers.

\subsection{Results}
Table~\ref{tab:summ_result} shows our evaluation results using ROUGE (F1) scores
and BERTScore on the Scriptbase corpus. Compared with the baseline models and previous work, our model achieves state-of-the-art results on all metrics. Specifically, our Select and Summarize model, which selects salient scenes, achieves 49.98, 12.11, and 47.95 on ROUGE-1/2/L scores and also shows improvements on BERTScore. Compared to a model which uses the full text of the movies, our model improves the performance by 3.83, 1.49, and 3.49 ROUGE-1/2/L points, respectively. The Lead-N baseline achieves better results than \citet{agarwal-etal-2022-creativesumm} with a ROUGE-1 of 17.69 for Lead-1024. Our model outperforms SUMM\textsuperscript{N} \cite{zhang-etal-2022-summn}, which can be attributed to better content selection using salient scenes compared to greedy content selection based on ROUGE.
As named entities and places are repeated across the movie script, the greedy alignment used in SUMM\textsuperscript{N} can result in false positives. 
Unlimiformer performance is low compared to our model and the two-stage model, possibly because it does not include explicit content selection.
The \citet{pu-etal-2022-two} model performs slightly better than using Full Text, as removing sentences based on heuristics allows it to include movie script text which would otherwise be truncated. FLAN-UL2 performs better than GPT-3.5-Turbo and FLAN-T5-XXL in a zero-shot setting but our fine-tuned model outperforms all three models. 

We also experimented with Pegasus-X \citep{phang2022investigating} instead of LED as the base summarization model for \textsc{Select \& Summ}. We found both models perform better when using our approach of selecting salient scenes compared to the full text, with LED demonstrating superior performance.

Figure~\ref{fig:length_dist}. also shows that our model yields improvements even though it uses only half the length (only salient scenes) of the original script. This demonstrates the effectiveness of salient scene selection in movie script summarization. Appendix~\ref{sec:generatedsumm} shows generated summaries for two movies.

\begin{table}[htbp]
\centering
\scalebox{0.9}{
\begin{tabular}{@{}lrrc@{}}
\toprule
 \textbf{Model} & \textbf{F1}        & \textbf{EM}  \\ \midrule
Full Text     & 22.70 & 13.81 \\
Two-Stage \citep{pu-etal-2022-two} & 25.21 & 14.35 \\
\textsc{Select \& Summ}      & \textbf{29.42} & \textbf{20.05}\\
\bottomrule
\end{tabular}}
\caption{Results of QAEval on summaries generated by Select and Summarize and baseline models.}
\label{tab:qa_eval}
\end{table}

\section{Automatic QA-based Evaluation}
Metrics like ROUGE (lexically based) and BERTScore (embedding based) are good for comparing the topic similarity between the reference and generated summaries, but fail to compare content-based factual consistency. To further evaluate the performance of our model, we used QAEval \citep{deutsch-etal-2021-towards}, a question-answering-based evaluation that generates question-answer pairs using the reference summaries. It then uses the model-generated summaries (candidates) to answer these questions, thereby measuring information overlap. It reports two standard answer verification methods used by SQuAD, F1 and exact match (EM) \citep{rajpurkar-etal-2016-squad}, averaged over all questions for all model-generated summaries. 

Before the final evaluation, we filtered the generated questions using a question filtering method similar to \citet{fabbri-etal-2022-qafacteval}, which is useful for removing spurious questions/answers (for example answers consisting of personal pronouns and wh-pronouns). Table~\ref{tab:qa_eval} shows results for QA\-Eval on summaries generated by models using full text input, the two-stage heuristic approach \citep{pu-etal-2022-two}, and Select and Summarize (our model). 
We find that Select and Summarize performs better in answering factual questions, with a mean F1 of 29.42 and a mean exact match of 20.05\%. Our model shows a clear improvement over using full movie scripts or a two-stage heuristic approach. 
\begin{table}[htbp]
\centering
\scalebox{0.9}{
\begin{tabular}{@{}lcccc@{}}
\toprule
\textbf{Model} & \textbf{R-1} & \textbf{R-2} & \textbf{R-L}& \textbf{\#P}\\\midrule
\textsc{Summ$^N$} & 32.48 & 5.85 & 27.55 & 400\\
\textsc{DialogLM} & \textbf{35.75} & 8.27 & 30.76 & 340\\ 
\textsc{SLED} & 35.20 & \textbf{8.70} & 19.40 & 406\\
\textsc{Select \& Summ} & 35.61 & 8.58 & \textbf{31.13} & 161\\
\bottomrule
\end{tabular}}
\caption{Zero-Shot performance of scene classifier on SummScreenFD compared with other baselines models. \#P is the number of fine-tuned parameters in millions.}
\label{tab:zero_shot}
\vspace{-2ex}
\end{table}

\section{Zero-Shot on SummScreen-FD}
We further investigate the performance of the scene saliency classifier on SummScreenFD \citep{chen-etal-2022-summscreen} as used in \textsc{Scrolls} benchmark \citep{shaham-etal-2022-scrolls}. SummScreenFD consists of transcripts of TV show episodes with human-written recaps. We performed a zero-shot classification of the salient scenes on the SummScreenFD and used only salient scenes to fine-tune LED for summarization. We compare the results with state-of-the-art methods on the dataset and report ROUGE scores. We observe that our model achieves comparable results to the state of the art on the SummScreenFD dataset as shown in Table \ref{tab:zero_shot}, but with fewer parameters \citep{10.1162/tacl_a_00547,zhong2022dialoglm}.
%
\section{Discussion and Conclusion}
In this paper, we introduced a dataset of 100 movies in which movie plot summaries are manually aligned with scenes in the corresponding movie script. Our dataset can be used to evaluate content selection strategies and extractive summarization for movie scripts. Using this dataset, we proposed a scene saliency classification model for the automatic identification of salient scenes in a movie script and introduced an abstractive summarization model that only uses the salient scenes to generate the movie summary. Our experiments showed that the proposed model achieves a significant improvement over the previous state of the art on the Scriptbase corpus for movie script summarization and performs comparable to the state of the art on the SummScreenFD dataset using zero-shot salient scene detection.

Our work demonstrates that the output of a summarization model can improve when content selection is performed (by using only the salient scenes). A good content selection strategy can in principle reduce the input size without compromising the quality of the generated output. As a result of the smaller input size, the computational and memory requirements of the underlying large language model can be significantly reduced.

\section*{Limitations}
Limitations of this work include that we defined the saliency of a scene as recall in user-written summaries. However, there are many aspects that can make a scene salient, including the presence of an important character or event in the scene, or just the fact that the scene is visually stunning. These factors can be explored in future work. Also, we discovered that many of the movie scripts in the Scriptbase corpus are not the final production scripts, which means they are different from the final movie as it was released. This imposes a limit on the quality of the summary that can be generated from a script. Our current model works in a pipeline of salient scene classification and then uses these scenes to summarize the movie. This means that it can propagate salience classification errors into the summarization step. Human evaluation of summaries generated from long-form text is challenging, as it requires human evaluators to read very long texts such as movie scripts. Therefore, future work is required to evaluate automatically generated movie summaries.

\section*{Ethics Statement}
\paragraph{Large Language Models:} This paper uses pre-trained large language models, which have been shown to be subject to a variety of biases, to occasionally generate toxic language, and to hallucinate content. Therefore, the summaries generated by our approach should not be released without automatic filtering or manual checking.

\paragraph{Experimental Participants:} The departmental ethics panel judged our human annotation study to be exempt from ethical approval, as all participants were employees of the University of Edinburgh, and as such were protected by employment law. Nevertheless, annotators were given a participant information sheet before they started work. They were also informed about the age rating of each movie script, based on which they could decide whether they want to annotate this script or not.
Participants were paid at the standard hourly rate for tutors and demonstrators at the university.

\section*{Acknowledgements}
This work was supported in part by the UKRI Centre for Doctoral Training in Natural Language Processing, funded by UK Research and Innovation (grant EP/S022481/1), Huawei, and the School of Informatics at the University of Edinburgh. We would like to thank the anonymous reviewers for their helpful feedback.

\bibliography{custom}

\appendix
\section{Further Implementation Details}

All experiments were performed on an A100 GPU with 80GB memory. It took approximately 22~hours to fully fine-tune the LED model and 30~hours for the Pegasus-X model. The LED-based models have 161M parameters, which were all fine-tuned. Our Scene Saliency Model has 60.2M parameters. The total number of parameters is 221.2M. The Pegasus-X has 568M parameters but its performance is lower than LED. 

For evaluation, we used Benjamin Heinzerling's implementation of Rouge\footnote{\url{https://github.com/bheinzerling/pyrouge}} and BERTScore with the microsoft/deberta-xlarge-mnli model.

\section{Scene Encoder Experiment}\label{sec:scene_embed}
\begin{table}[htbp]
\centering
\begin{tabular}{@{}lccc@{}} 
\toprule
\textbf{Model} & \textbf{P} & \textbf{R} & \textbf{F1} \\ 
\midrule
BART & 66.13 & 66.48 & 66.06  \\
LED (Encoder) & 67.18 & 63.62 & 64.11 \\
\textbf{Roberta} & \textbf{68.38} & \textbf{68.13} & \textbf{68.01}   \\
\bottomrule
\end{tabular}
\caption{Performance of Scene Saliency Model for different base models as scene encoder.}
\label{tab:scene_embed_tab}
\end{table}

We compared the performance of Roberta with that of BART \citep{lewis-etal-2020-bart} and LED (Encoder only) as the base models for computing scene embeddings in the classification of salient scenes. For each model, we employed the large variant and extracted the encoder's last hidden state as scene embeddings. We report the results of scene saliency classification with different base models in Table~\ref{tab:scene_embed_tab}. Among these models, Roberta's embeddings performed marginally better and also had fewer parameters.

\begin{table}[htbp]
\centering
\begin{tabular}{@{}lll@{}}  
  \hline
\textbf{Mean Precision} & \textbf{Mean Recall} & \textbf{Mean F1} \\ 
  \hline
  $68.09 \pm 0.006$ & $68.03 \pm 0.005$ & $67.70 \pm 0.003$ \\ 
\toprule
\end{tabular}
\caption{Cross validation result for scene saliency classifier.}
\label{tab:append_class}
\end{table}

\section{Classifier Robustness}\label{sec:robustness}
To study the robustness of the scene saliency classifier we performed k-fold cross-validation with $k=5$. We report mean results with standard deviation across all folds in Table~\ref{tab:append_class}. The low standard deviation shows that the performance of the scene classifier is robust across different folds.

\begin{table}[htbp]
\centering
\scalebox{0.8}{
\begin{tabular}{@{}lccc@{}}
\toprule
\textbf{Metric} & \textbf{Two-Stage Heuristic} & \textbf{SELECT \& SUMM (LED)} \\ \midrule
\textbf{R-1}     & $46.89$ ($44.17$-$47.20$)            & $49.98$ ($48.75$-$50.84$)              \\
\textbf{R-2}     & $11.11$ ($9.91$-$11.68$)              & $12.11$ ($11.71$-$12.87$)              \\
\textbf{R-L}     & $44.65$ ($42.35$-$46.20$)             & $47.95$ ($47.19$-$49.29$)              \\ 
\bottomrule
\end{tabular}
}
\caption{Performance of Scene Saliency Model for different base models as scene encoder.}
\label{tab:summ_stats_tab}
\end{table}
\section{Statistics for Summarization Result}\label{sec:summ_stats}
All the ROUGE scores reported in the paper are mean F1 scores with bootstrap resampling with 1000 number of samples.
To assess the significance of the results, we are reporting 95\% confidence interval results for our model and the closest baseline in Table~\ref{tab:summ_stats_tab}.

\section{Samples of Movie Summaries}\label{sec:generatedsumm}

\begin{table*}[!htbp]
\small
\begin{tabularx}{\linewidth}{X}
\toprule
\textbf{Movie: Lincoln}\\
\midrule
\textbf{Gold Summary}\\

In January 1865, United States President Abraham Lincoln expects the Civil War to end soon, with the defeat of the Confederate States. He is concerned that his 1863 Emancipation Proclamation may be discarded by the courts after the war and that the proposed Thirteenth Amendment will be defeated by the returning slave states. He feels it imperative to pass the amendment beforehand, to remove any possibility that freed slaves might be re-enslaved. The Radical Republicans fear the amendment will be defeated by some who wish to delay its passage; support from Republicans in the border states is not yet assured. The amendment also requires the support of several Democratic congressmen to pass. With dozens of Democrats being lame ducks after losing their re-election campaigns in the fall of 1864, some of Lincoln's advisors believe he should wait for a new Republican-heavy Congress. Lincoln remains adamant about having the amendment in place before the war is concluded and the southern states are re-admitted. Lincoln's hopes rely upon Francis Preston Blair, a founder of the Republican Party whose influence could win over members of the border state conservative faction. With Union victory in the Civil War highly likely but not yet secured, and with two sons serving in the Union Army, Blair is keen to end hostilities quickly before the spring thaw arrives and the armies march again. Therefore, in return for his support, Blair insists that Lincoln allow him to engage the Confederate government in peace negotiations. However, Lincoln knows that significant support for the amendment comes from Radical Republicans, for whom negotiated peace is unacceptable. Unable to proceed without Blair's support, Lincoln reluctantly authorizes Blair's mission. In the meantime, Lincoln and Secretary of State William Seward work to secure Democratic votes for the amendment. Lincoln suggests they concentrate on the lame-duck Democrats, as they will feel freer to vote as they choose and soon need employment; Lincoln will have many federal jobs to fill as he begins his second term. Though Lincoln and Seward are unwilling to offer monetary bribes to the Democrats, they authorize agents to contact Democratic congressmen with offers of federal jobs in exchange for their support. Meanwhile, Lincoln's son, Robert, returns from law school and announces his intention to discontinue his studies and enlist in the Union Army, hoping to earn a measure of honor and respect outside of his father's shadow before the war's end. Lincoln reluctantly secures an officer's commission for Robert. The First Lady is aghast, fearing that he will be killed. She furiously presses her husband to pass the amendment and end the war, promising woe upon him if he should fail. At a critical moment in the debate in the House of Representatives, racial-equality advocate Thaddeus Stevens agrees to moderate his position and argue that the amendment represents only legal equality, not a declaration of actual equality. Meanwhile, Confederate envoys are ready to meet with Lincoln to discuss terms for peace, but he instructs they be kept out of Washington as the amendment approaches a vote on the House floor. Rumor of their mission circulates, prompting both Democrats and conservative Republicans to advocate postponing the vote. In a carefully worded statement, Lincoln denies there are envoys in Washington, and the vote proceeds, passing by a margin of just two votes. Black visitors to the gallery celebrate, and Stevens returns home to his "housekeeper" and lover, a black woman. When Lincoln meets with the Confederates, he tells them slavery cannot be restored, as the North is united for ratification of the amendment, and several of the southern states' reconstructed legislatures would also vote to ratify. As a result, the peace negotiations fail, and the war continues. On April 3, Lincoln visits the battlefield at Petersburg, Virginia, where he exchanges a few words with Lieutenant General Ulysses S. Grant. On April 9, Grant receives General Robert E. Lee's surrender at Appomattox Courthouse. On April 14, a cheerful Lincoln expresses to his wife that they will be happy in the future and later meets members of his cabinet to discuss future measures to enfranchise blacks, before leaving for Ford's Theatre. That night, while Lincoln's son Tad is watching Aladdin and the Wonderful Lamp at Grover's Theatre, the manager suddenly stops the play to announce that the President has been shot. The next morning, at the Petersen House, Lincoln dies with a peaceful expression across his face; in a flashback, Lincoln finishes his second inaugural address on March 4.\\
\midrule
\textbf{Sample of Question-Answer pairs for Evaluation}\\
\midrule
\begin{tabular}{p{0.72\textwidth} p{0.2\textwidth}}
What city are Confederate envoys ready to meet with Lincoln to discuss terms for peace? & \textcolor{correct}{Washington}\\\\
As a result the peace negotiations fail, what continues? &  \textcolor{correct}{war}\\\\
On April 3, who visits the battlefield at Petersburg, Virginia, where he exchanges a few words with Lieutenant General Ulysses S. Grant? & \textcolor{correct}{Lincoln}\\\\
Meanwhile, Confederate envoys are ready to meet with Lincoln to discuss terms for peace, but he instructs they be kept out of Washington as what approaches a vote on the House floor? & \textcolor{correct}{the amendment}\\\\
What are Confederate envoys ready to meet with Lincoln to discuss terms for, but he instructs they be kept out of Washington as the amendment approaches a vote on the House floor? & \textcolor{correct}{peace}\\\\
In a carefully worded statement, Lincoln denies there are what in Washington? &\textcolor{correct}{envoys}\\
\end{tabular}\\
\bottomrule
\caption{Gold reference summary for the movie Lincoln with sample of question-answer pairs generated for evaluation.}
\label{tab:prompt_outline_item}
\end{tabularx}
\end{table*}

\begin{table*}[!htbp]
\small
\begin{tabularx}{\linewidth}{X}
\toprule
\textbf{Movie: Lincoln}\\
\midrule
\textbf{Generated Summary}\\
 In 1863, U.S. President Abraham Lincoln and his Cabinet prepare for war against the Confederate States of America. Secretary of War Edwin M. Stanton issues a war proclamation declaring slaves in the Confederate States to be property confiscated during the American Civil War. Drawing on his war powers, Lincoln declares that slaves in the thirteen southern states are property of the Union, and that he can confiscate them and free them with his 1863 Emancipation Proclamation. This, however, is challenged by the states' refusal to ratify the Thirteenth Amendment to the United States' Constitution, which would end slavery. The courts refuse to recognize the document, which would ban slavery, forcing Lincoln to act by proclamation. He reluctantly authorizes the Emancipation Proclamation, but suspends its implementation pending a vote in Congress. The Southern states refuse to ratify the amendment, preventing it from being put before a vote until they are readmitted to the Union. Only two years later, on Good Friday, Lincoln takes his wife, Mary Todd Lincoln, out for a buggy ride. They talk about plans to travel to the Holy Land and how much they want to get out of Washington, D.C. Before returning to the White House, they view a picture of Tad Lincoln, Abraham Lincoln's son, with his mistress, Mary Keckley. Mary reveals that she was a slave, and that her husband was killed by their son. Lincoln is deeply disturbed by this revelation. A year later, on April 4, 1865, the southern states are readmitted to the Union, but they are not readmitted to block ratification of the Thirteenth Amendment. Secretary of the Navy Gideon Welles issues a report that the southern states have refused to accede to the amendment, preventing it from being ratified. This leads to friction between Lincoln's pro-slavery advisers and the pro-slavery delegation in Congress. They attempt to convince the reluctant Wood, Baird, and Blair to delay a vote on the amendment in the House until they have information from the President on the whereabouts of the Confederate States' commissioners, who are supposed to have arrived in Washington, D.C., to negotiate a solution to end the war. The White House attempts to find evidence of the commissioners' arrival, but they rely on affidavits from loyal citizens recently returned from Richmond, Virginia. Meanwhile, Lincoln's popularity continues to rise, and he continues to insist on action on the war. Privately, he expresses his hope that the war will end soon and they can begin a new life together, but his Vice President, Jefferson Davis, persuades him to delay the vote on the amendment, hoping to secure enough votes in the Senate to block ratification. When the Senate fails to act on the amendment, and the House refuses to act, Lincoln is forced to take action, forcing a shutdown of the government. Seward urges Lincoln to declare the war over and negotiate a peace, but Lincoln refuses. Seward pressures him to declare a truce, which would allow the states to resume their former position in the Union and block ratification of the Thirteenth Amendment. On April 5, 1865, the war ends at Appomattox Courthouse in Appomattox, Virginia. Confederate General Robert E. Lee surrenders to General Ulysses S. Grant at Appomattox. A great deal of animosity remains between Lincoln and Lee, and the two men ride together to the McLean House, where the wounded are brought to view the final Confederate corpses from the Appomattox Massacre. When Lee approaches the McLeans, Grant stops him, takes his hat, and salutes him. Lee is visibly moved by this gesture of respect. The film ends with Lincoln and Mary walking arm in arm to the edge of the balcony of the McLean House, looking down on the battlefield.\\
\midrule
\textbf{Answers using generated summary}\\
\midrule
\begin{tabular}{p{0.72\textwidth} p{0.2\textwidth}}
What city are Confederate envoys ready to meet with Lincoln to discuss terms for peace ? & \textcolor{correct}{Washington}\\\\
As a result the peace negotiations fail, what continues? & \textcolor{correct}{war}\\\\
On April 3, who visits the battlefield at Petersburg, Virginia, where he exchanges a few words with Lieutenant General Ulysses S. Grant?& \textcolor{correct}{Lincoln}\\\\
Meanwhile, Confederate envoys are ready to meet with Lincoln to discuss terms for peace, but he instructs they be kept out of Washington as what approaches a vote on the House floor? & \textcolor{red}{Thirteenth} \textcolor{correct}{Amendment}\\ \\
What are Confederate envoys ready to meet with Lincoln to discuss terms for, but he instructs they be kept out of Washington as the amendment approaches a vote on the House floor? & \textcolor{red}{end the war}\\\\
In a carefully worded statement, Lincoln denies there are what in Washington? & \textcolor{red}{commissioners}\\
\end{tabular}\\
\bottomrule
\caption{Model generated summary of the movie Lincoln with answers to the generated question. The correct answers are represented by the green color, while the incorrect answers are represented by the red color. Some answers can be partially correct which can have both the colors.}
\label{tab:prompt_outline_item}
\end{tabularx}
\end{table*}

\begin{table*}[!htbp]
\small
\begin{tabularx}{\linewidth}{X}
\toprule
\textbf{Movie: Black Panther}\\
\midrule
\textbf{Gold Summary}\\
Thousands of years ago, five African tribes war over a meteorite containing the metal vibranium. One warrior ingests a "heart-shaped herb" affected by the metal and gains superhuman abilities, becoming the first "Black Panther". He unites all but the Jabari Tribe to form the nation of Wakanda. Over centuries, the Wakandans use the vibranium to develop advanced technology and isolate themselves from the world by posing as a Third World country. In 1992, Wakanda king T'Chaka visits his brother N'Jobu, who is working undercover in Oakland, California. T'Chaka accuses N'Jobu of assisting black-market arms dealer Ulysses Klaue with stealing vibranium from Wakanda. N'Jobu's partner reveals he is Zuri, another undercover Wakandan, and confirms T'Chaka's suspicions. In the present day, following T'Chaka's death, his son T'Challa returns to Wakanda to assume the throne. He and Okoye, the leader of the Dora Milaje regiment, extract T'Challa's ex-lover Nakia from an undercover assignment so she can attend his coronation ceremony with his mother Ramonda and younger sister Shuri. At the ceremony, the Jabari Tribe's leader M'Baku challenges T'Challa for the crown in ritual combat. T'Challa defeats M'Baku and persuades him to yield rather than die. When Klaue and his accomplice Erik Stevens steal a Wakandan artifact from a London museum, T'Challa's friend and Okoye's lover W'Kabi urges him to bring Klaue back alive. T'Challa, Okoye, and Nakia travel to Busan, South Korea, where Klaue plans to sell the artifact to CIA agent Everett K. Ross. A firefight erupts, and Klaue attempts to flee but is caught by T'Challa, who reluctantly releases him to Ross' custody. Klaue tells Ross that Wakanda's international image is a front for a technologically advanced civilization. Erik attacks and extracts Klaue as Ross is gravely injured protecting Nakia. Rather than pursue Klaue, T'Challa takes Ross to Wakanda, where their technology can save him. While Shuri heals Ross, T'Challa confronts Zuri about N'Jobu. Zuri explains that N'Jobu planned to share Wakanda's technology with people of African descent around the world to help them conquer their oppressors. As T'Chaka arrested N'Jobu, the latter attacked Zuri and forced T'Chaka to kill him. T'Chaka ordered Zuri to lie that N'Jobu had disappeared and left behind N'Jobu's American son to maintain the lie. This boy grew up to be Stevens, a black ops U.S. Navy SEAL who adopted the name "Killmonger". Meanwhile, Killmonger kills Klaue and takes his body to Wakanda. He is brought before the tribal elders, revealing his identity to be N'Jadaka and stating his claim to the throne. Killmonger challenges T'Challa to ritual combat, where he kills Zuri, defeats T'Challa, and hurls him over a waterfall to his presumed death. Killmonger ingests the heart-shaped herb and orders the rest incinerated, but Nakia extracts one first. Killmonger, supported by W'Kabi and his army, prepares to distribute shipments of Wakandan weapons to operatives around the world. Nakia, Shuri, Ramonda, and Ross flee to the Jabari Tribe for aid. They find a comatose T'Challa, rescued by the Jabari in repayment for sparing M'Baku's life. Healed by Nakia's herb, T'Challa returns to fight Killmonger, who dons his own nanotech suit, similar to T'Challa's. W'Kabi and his army fight Shuri, Nakia, and the Dora Milaje, while Ross remotely pilots a jet and shoots down planes carrying the vibranium weapons. M'Baku and the Jabari arrive to reinforce T'Challa. Confronted by Okoye, W'Kabi and his army stand down. Fighting in Wakanda's vibranium mine, T'Challa disrupts Killmonger's suit and stabs him. Killmonger refuses to be healed, choosing to die a free man rather than be incarcerated; T'Challa takes him to the waterfall where they fought, where Killmonger dies peacefully. T'Challa establishes an outreach center at the building where N'Jobu died, to be run by Nakia and Shuri. In a mid-credits scene, T'Challa appears before the United Nations to reveal Wakanda's true nature to the world.\\
\midrule
\textbf{Sample of Question-Answer pairs for Evaluation}\\
\midrule
\begin{tabular}{ll}
Thousands of years ago, five African tribes war over a meteorite containing what metal? & \color{correct}vibranium\\\\
Where do T'Challa and Okoye travel to South Korea where Klaue plans to sell the artifact? & \color{correct}Busan\\\\
What does one warrior ingest affected by the metal and gain superhuman abilities? & \color{correct}heart-shaped herb\\\\
Wakandans use the vibranium to develop advanced technology and isolate themselves from what? & \color{correct}the world\\\\
Who do T'Challa, Okoye, and Nakia travel to Busan to sell the artifact to? & \color{correct}Everett K. Ross\\\\
While Shuri heals Ross, who does T'Challa confront about N'Jobu? & \color{correct}Zuri\\
\end{tabular}\\
\bottomrule
\caption{Gold reference summary for the movie Black Panther with sample of question-answer pairs generated for evaluation.}
\label{tab:prompt_outline_item}
\end{tabularx}
\end{table*}

\begin{table*}[!htbp]
\small
\begin{tabularx}{\linewidth}{X}
\toprule

\textbf{Movie: Black Panther}\\
\midrule
\textbf{Generated Summary}
After the death of his father King T'Chaka at the United Nations, Prince T'Challa of the African country of Wakanda adopts the title "Black Panther" and assumes the name T'Challa. A meteorite lands on Africa, striking the continent of Africa, and five tribes settle on it including Wakanda. The native people of Wakanda named Wakanda "Wakanda". The people live in harmony with each other and use vibranium, a mineral that gives them superhuman strength, speed, and other powers. A warrior from Wakanda named N'Jobu uses vibranium to make the Heart Shaped Herb, which gives his son T'Challa his powers. A CIA agent, Eric Stevens, traces the Heart Shaped Herb to Ulysses Klaue, a black marketeer attempting to obtain vibranium. T'Challa's aunt and uncle, Nakia and Okoye, lead a mission to retrieve vibranium from Klaue at the request of the African Dora Milaje. They capture Klaue and bring him to T'Challa at the Great Mound. T'Challa challenges Klaue for the throne of Wakanda. The meeting is interrupted by the arrival of Young T'Chaka, T'Challa's half-brother and the son of King T'Chaka. T'Challa takes Klaue on a train trip back to Wakanda with the Dora Milaje, but Klaue injures Ross, a CIA agent. T'Challa and Ross confront Klaue at the train station, and Klaue declares his intention to sell the vibranium to an American buyer in South Korea. T'Challa's aunt and uncle, Nakia and Okoye, tell him that if he sells the vibranium, the Dora Milaje will recover the money and return it to Wakanda. T'Challa agrees to this, and they fly to South Korea to retrieve the Heart Shaped Herb. At the U.S. embassy in Busan, South Korea, they learn that Klaue is actually an operative of the CIA, and that the Heart Shaped Herb has been smuggled into Wakanda by an unknown party. T'Challa and Ross confront Klaue in a casino. Klaue threatens to kill Ross if T'Challa does not hand over the vibranium. T'Challa offers Ross a deal: if he delivers the vibranium, he will be allowed to return to Wakanda. Ross refuses, and Klaue attempts to kill T'Challa, but is stopped by Young T'Challa. T'Challa heals Ross, and the two embark to return to Wakanda. Along the way, Ross reveals himself to T'Challa as an operative of the CIA, and T'Challa takes him back to the CIA. T'Challa and Nakia arrive in Wakanda, where they meet Nakia's mother, Queen Mother Ramonda, and the Dora Milaje. They purchase vibranium from a merchant at the port of Busan. T'Challa presents the Heart Shaped Herb to the people of Wakanda, and Nakia gives birth to a son, T'Challa born T'Chaka. T'Challa and Nakia take the Heart Shaped Herb to Jabariland, where it is discovered that Nakia is a spy for the CIA. The Heart Shaped Herb gives T'Challa superhuman strength and other powers. He goes to Jabariland to find Nakia and finds her serving as a midwife to a dying M'Baku. M'Baku reveals himself to be a spy sent by the CIA to find T'Challa and is confronted by T'Challa. A gunfight ensues, in which T'Challa gains the upper hand, but M'Baku seizes the vibranium and attempts to kill T'Challa. T'Challa uses the Heart Shaped Herb to keep M'Baku at bay, and returns to Wakanda with Nakia, Ross, and the Heart Shaped Herb to deliver T'Challa to his people. Nakia betrays T'Challa to the CIA, and Ross is sent to retrieve the vibranium. T'Challa and Nakia escape with the Heart Shaped Herb and take it to Jabariland, where it is recovered by W'Kabi, the leader of the Jabari tribe. The tribe's leader, W'Kabi, uses it to heal T'Challa, and he returns to Wakanda with T'Challa and the Heart Shaped Herb to reunite the tribes. In the present, T'Challa and Nakia celebrate the opening of Wakanda's new center for science and information exchange with the rest of the world.\\
\midrule
\textbf{Answers using generated summary}\\
\midrule
\begin{tabular}{ll}
Thousands of years ago, five African tribes war over a meteorite containing what metal? & \color{correct}{vibranium}\\\\
Where do T'Challa and Okoye travel to South Korea where Klaue plans to sell the artifact? & \color{correct}Busan\\\\
What does one warrior ingest affected by the metal and gain superhuman abilities? & \textcolor{red}{heart shaped} \color{correct}herb\\\\
Wakandans use the vibranium to develop advanced technology and isolate themselves from what? & \textcolor{red}{United Nations}\\\\
Who do T'Challa, Okoye, and Nakia travel to Busan to sell the artifact to? & \textcolor{red}{American buyer}\\\\
While Shuri heals Ross, who does T'Challa confront about N'Jobu? & \textcolor{red}{M'Baku}\\
\end{tabular}\\
\bottomrule
\caption{Model generated summary of the movie Black Panther with answers to the generated question. The correct answers are represented by the green color, while the incorrect answers are represented by the red color. Some answers can be partially correct which can have both the colors.}
\label{tab:prompt_outline_item}
\end{tabularx}
\end{table*}

\end{document}